%% file: main.tex
\documentclass[11pt,a4paper]{article}
\usepackage[hyperref]{emnlp2018}
\usepackage{times}
\usepackage{latexsym}
\usepackage{subcaption}

\usepackage{url}

\usepackage{color}
\usepackage{graphicx}
\usepackage{import}
\usepackage{amsmath, amssymb, amsthm} 
\newcommand{\red}[1]{{\color{red}{#1}}}
\newcommand{\blue}[1]{{\color{blue}{#1}}}
\newcommand{\green}[1]{{\color{green!70!black!90}{#1}}}

\newtheorem{ex}{Example}
\usepackage{enumitem}
\usepackage{booktabs, hhline}
\usepackage{multirow}

\usepackage[textsize=tiny]{todonotes}

\usepackage{linguex}

\aclfinalcopy 
\def\aclpaperid{43} 

\title{Analysing the potential of seq-to-seq models\\ for incremental interpretation in task-oriented dialogue}
\author{Dieuwke Hupkes, Sanne Bouwmeester, \and Raquel Fern\'{a}ndez\\
	Institute for Logic Language and Computation\\
    University of Amsterdam\\
    {\tt \{d.hupkes,raquel.fernandez\}@uva.nl}\\
    {\tt sanne.bouwmeester1@student.uva.nl}
    }

\date{}

\begin{document}
\maketitle

\begin{abstract}
We investigate how encoder-decoder models trained on a synthetic  dataset of task-oriented dialogues process disfluencies, such as hesitations and self-corrections. We find that, contrary to earlier results, disfluencies have very little impact on the task success of seq-to-seq models with attention. Using visualisations and diagnostic classifiers, we analyse the representations that are incrementally built by the model, and discover that models develop little to no awareness of the structure of disfluencies. However, adding disfluencies to the data appears to help the model create clearer representations overall, as evidenced by the attention patterns the different models exhibit.
\end{abstract}


\input{introduction}

\input{babi}

\input{setup}

\input{analysis}\label{sec:analysis}

\input{attention}

\input{analysis2}\label{sec:analysis_ling}

\input{discussion}


\bibliographystyle{acl_natbib_nourl}
\bibliography{ref}

\end{document}

%% file: introduction.tex
\section{Introduction}

The use of Recurrent Neural Networks (RNNs) to tackle sequential language tasks has become standard in natural language processing, after impressive accomplishments in speech recognition, machine translation, and entailment \cite[e.g.,][]{sutskever2014sequence,bahdanau2015attention,kalchbrenner2014convolutional}. 
Recently, RNNs have also been exploited as tools to model dialogue systems. Inspired by neural machine translation, researchers such as \citet{ritter} and \citet{vinyals2015neural} pioneered an approach to open-domain chit-chat conversation based on sequence-to-sequence models \citep{sutskever2014sequence}. 
In this paper, we focus on task-oriented dialogue, where the conversation serves to fulfil an independent goal in a given domain. 
Current neural dialogue models for task-oriented dialogue tend to equip systems with external memory components \citep{bordes2016learning}, since key information needs to be stored for potentially long time spans. 
One of our goals here is to analyse to what extent sequence-to-sequence models without external memory can deal with this challenge.  

In addition, we consider language realisations that include disfluencies common in dialogue interaction, such as repetitions and self-corrections (e.g., {\em I'd like to make a reservation for six, I mean, for eight people}). Disfluencies have been investigated extensively in psycholinguistics, with a range of studies showing that they affect sentence processing in intricate ways \citep{levelt1983monitoring,tree1995effects,bailey2003disfluencies,FerreiraBailey2004,LauFerreira2005,brennan2001listeners}. Most computational work on disfluencies, however, has focused on detection rather than on disfluency processing and interpretation \cite[e.g.,][]{stolcke1996statistical,heeman1999speech,zwarts2010detecting,qian2013disfluency,HoughPurver2014,HoughSchlangen2017}. In contrast, our aim is to get a better understanding of how RNNs process disfluent utterances and to analyse the impact of such disfluencies on a downstream task---in this case, issuing an API request reflecting the preferences of the user in a task-oriented dialogue. 

For our experiments, we use the synthetic dataset bAbI \citep{bordes2016learning} and a modified version of it called bAbI+ which includes disfluencies \citep{shalyminov2017challenging}. The dataset contains simple dialogues between a user and a system in the restaurant reservation domain, which terminate with the system issuing an API call that encodes the user's request. In bAbI+, disfluencies are probabilistically inserted into user turns, following distributions in human data.  Thus, while the data is artificial and certainly simplistic, its goal-oriented nature offers a rare opportunity: by assessing whether the system issues the right API call, we can study, in a controlled way, whether and how the model builds up a relevant semantic/pragmatic interpretation when processing a disfluent utterance---a key aspect that would not be available with unannotated natural data.

%% file: babi.tex
\section{Data}
\label{sec:babi}

In this section, we discuss the two datasets we use for our experiments: bAbI \citep{bordes2016learning} and bAbI+ \citep{shalyminov2017challenging}.

\subsection{bAbI}
The bAbI dataset consists of a series of synthetic dialogues in English, representing human-computer interactions in the context of restaurant reservations.
The data is broken down  into six subtasks that individuate different abilities that dialogue systems should have to conduct a successful conversation with a human.
We focus on Task 1, which tests the capacity of a system to ask the right questions and integrate the answers of the user to issue an API call that matches the user's preferences regarding four semantic slots: cuisine, location, price range, and party size. A sample dialogue can be found in example \ref{ex:repair}, Section~\ref{sec:error}.

\paragraph{Data} 
The training data for Task 1 is deliberatively kept simple and small, consisting of 1000 dialogues with on average 5 user and 7 system utterances.
An additional 1000 dialogues based on different user queries are available for validation and testing, respectively.
The overall vocabulary contains 86 distinct words. There are 7 distinct system utterances and 300 possible API calls.

\paragraph{Baselines}
 Together with the dataset, \citet{bordes2016learning} present several baseline models for the task. All the methods proposed are retrieval based, i.e., the models are trained to select the best system response from a set of candidate responses (in contrast to the models we investigate in the present work, which are generative---see Section~\ref{sec:results}). 
The baseline models include classical information retrieval (IR) methods such as TF-IDF and nearest neighbour approaches, as well as an end-to-end recurrent neural network.
 \citeauthor{bordes2016learning} demonstrate that the end-to-end recurrent architecture---a memory network \citep{sukhbaatar2015memory}---outperforms the classical IR methods as well as supervised embeddings, obtaining a 100\% accuracy on retrieving the correct API calls.

\subsection{bAbI+}
\label{sec:babi+}

\citet{shalyminov2017challenging} observe that the original bAbI data lack naturalness and variation common in actual dialogue interaction. To introduce such variation while keeping lexical variation constant, they insert speech disfluencies, using a fixed set of templates that are probabilistically applied to the user turns of the original bAbI Task 1 dataset. In particular, three types of disfluencies are introduced: hesitations \ref{ex:hesitation}, restarts \ref{ex:restart}, and self-corrections \ref{ex:correction}, in around 21\%, 40\% and 5\% of the user's turns, respectively.\footnote{The inserted material is in italics in the examples.}

\ex. We will be \textit{uhm} eight 
\label{ex:hesitation}

\ex. Good morning \textit{uhm yeah good morning}
\label{ex:restart}

\ex. I would like \textit{a French uhm sorry} a Vietnamese restaurant
\label{ex:correction}

\noindent
\citet{eshghi2017bootstrapping} use the bAbI+ dataset to show that a grammar-based semantic parser specifically designed to process incremental dialogue phenomena is able to handle the bAbI+ data without having been directly exposed to it, achieving 100\% accuracy on API-call prediction. They then investigate whether the memory network approach by \citet{bordes2016learning} is able to generalise to the disfluent data, finding that the model obtains very poor accuracy (28\%) on API-call prediction when trained on the original bAbI dataset and tested on bAbI+. 
\citet{shalyminov2017challenging} further show that, even when the model is explicitly trained on bAbI+, its performance decreases significantly, achieving only 53\% accuracy. 

This result, together with the high level of control on types and frequency of disfluencies offered by the bAbI+ scripts, makes the bAbI+ data an excellent testbed for studying the processing of disfluencies by recurrent neural networks.

%% file: setup.tex
\section{Generative bAbI+ Modelling}
\label{sec:results}

\begin{table*}[!ht]\centering
\begin{tabular}{|l@{\ }c@{\ }l||cc|cc||c|}\hline
 		& & & \multicolumn{2}{c|}{\bf seq2seq} &  \multicolumn{2}{c||}{\bf attentive seq2seq} & MemN2N \\\hline
 \multicolumn{3}{|c||}{train / test}      & utterances & API calls & utterances & API calls & API calls \\\hline
bAbI   & /  & bAbI     &  100 (100)   & 0.02 (66.4) &  100 (100)  &   100 (100) & 100\\ 
bAbI+ & /  & bAbI+   &  100 (100)  & 0.2 (80.6) & 100 (100)   & 98.7 (99.7) & 53 \\
bAbI   & /  & bAbI+   & 81.4 (83.3)  & 0.00 (58.2) & 91.5 (92.8)  & 50.4 (90.1) & 28 \\
bAbI+ & /  & bAbI     &  100 (100)  & 0.2 (81.4) & 100 (100)    &  99.2 (100)       & 99 \\ \hline
\end{tabular}
\caption{Sequence accuracy (word accuracy in brackets) on the test set for utterances (non-API call responses) and API calls only. The last column shows accuracy on the test set for the retrieval-based memory-network system, as reported by \newcite{shalyminov2017challenging} .}
\label{tab:results}
\end{table*}

We start with replicating the results of \citet{shalyminov2017challenging} and \citet{eshghi2017bootstrapping} using a \textit{generative} rather than retrieval based model.
For this replication, we use a vanilla one-layer encoder-decoder model \citep{sutskever2014sequence} without any external memory.
We train models with and without an attention mechanism \citep{bahdanau2015attention} and compare their results.
We perform a modest grid search over hidden layer and embedding sizes and find that an embedding size of 128 and a hidden layer size of 500 appear to be minimally required to achieve a good performance on the task.
We therefore fix the embedding and hidden layer size to 128 and 500, respectively, for all further experiments.

\subsection{Training}
All models are trained to predict the system utterances of all of the 1000 training dialogues of the bAbI and bAbI+ dataset, respectively, including the final API call. 
After each user turn, models are asked to generate the next system utterance in the dialogue, given the dialogue history up to that point, which consists of all human and system utterances that previously occurred in that dialogue. 
The model's parameters are updated using stochastic gradient descent on a cross-entropy loss (using mini-batch size 32), with Adam \citep{kingma2014adam} as optimiser (learning rate 0.001). 
All models are trained until convergence, which was reached after $\sim$20 epochs.

\subsection{Evaluation}
Following \citet{shalyminov2017challenging}, we use a 2$\times$2 paradigm in which we train models either on bAbI or bAbI+ data and evaluate their performance on the test set of the same dataset, as well as across datasets.
We report both the percentage of correct words in the generated responses (\textit{word accuracy}) and the percentage of responses that were entirely correct (\textit{sequence accuracy}). 
Additionally, we separately report the word and sequence accuracy of the API calls generated at the end of each dialogue.
Note that these metrics are more challenging than the retrieval-based ones used by \newcite{bordes2016learning} and \newcite{eshghi2017bootstrapping}, as the correct response has to be generated word by word, rather than merely being selected from a set of already available candidate utterances.

\subsection{Results}
Our results can be found in Table \ref{tab:results}.  
The results obtained with the bAbI/bAbI and bAbI+/bAbI+ conditions indicate that an encoder-decoder model with attention can achieve near-perfect accuracy on Task 1 (predicting the right API call), whereas a model without attention cannot (sequence accuracy for API calls is only 0.02\% on  bAbI/bAbI and 0.2\% on bAbI+/bAbI+).
This suggests that, in line with what was posed by \citet{bordes2016learning}, the bAbI Task 1 requires some form of memory that goes beyond what is available in a vanilla sequence-to-sequence model. To solve the task, however, using an attention mechanism suffices---a more complicated memory such as present in memory networks is not necessary.

Furthermore, our results confirm that models trained on data without disfluencies struggle to generalise when these are introduced at testing time (bAbI/bAbI+): While the overall accuracy of the dialogue is still high (91.5\% of utterances are correct), API call accuracy falls back to 50.4\%.
Models trained on data containing disfluencies, however, show near-perfect accuracy on disfluent test data (98.7\% on bAbI+/bAbI+)---a result that stands in stark contrast with the findings of \citet{eshghi2017bootstrapping} and \citet{shalyminov2017challenging}.

%% file: analysis.tex
\section{Generalisation to Disfluent Data}
\label{sec:gen}

In this section, we analyse the potential for generalisation of the encoder-decoder model with attention by focusing on the bAbI/bAbI+ condition, where the model trained on bAbI data is tested on bAbI+. 
As shown in Table~\ref{tab:results}, while the model performs perfectly on the bAbI corpus, it achieves only $\sim$50\% accuracy on API call prediction when it is asked to generalised to bAbI+ data.
Here we aim to shed light on these results by studying the errors made by the model and visualising the patterns of the attention component of the network.

\subsection{Qualitative error analysis}
\label{sec:error}

\begin{figure*}[!h]
\includegraphics[trim=25mm 111mm 82mm 92mm, clip, width=\linewidth]{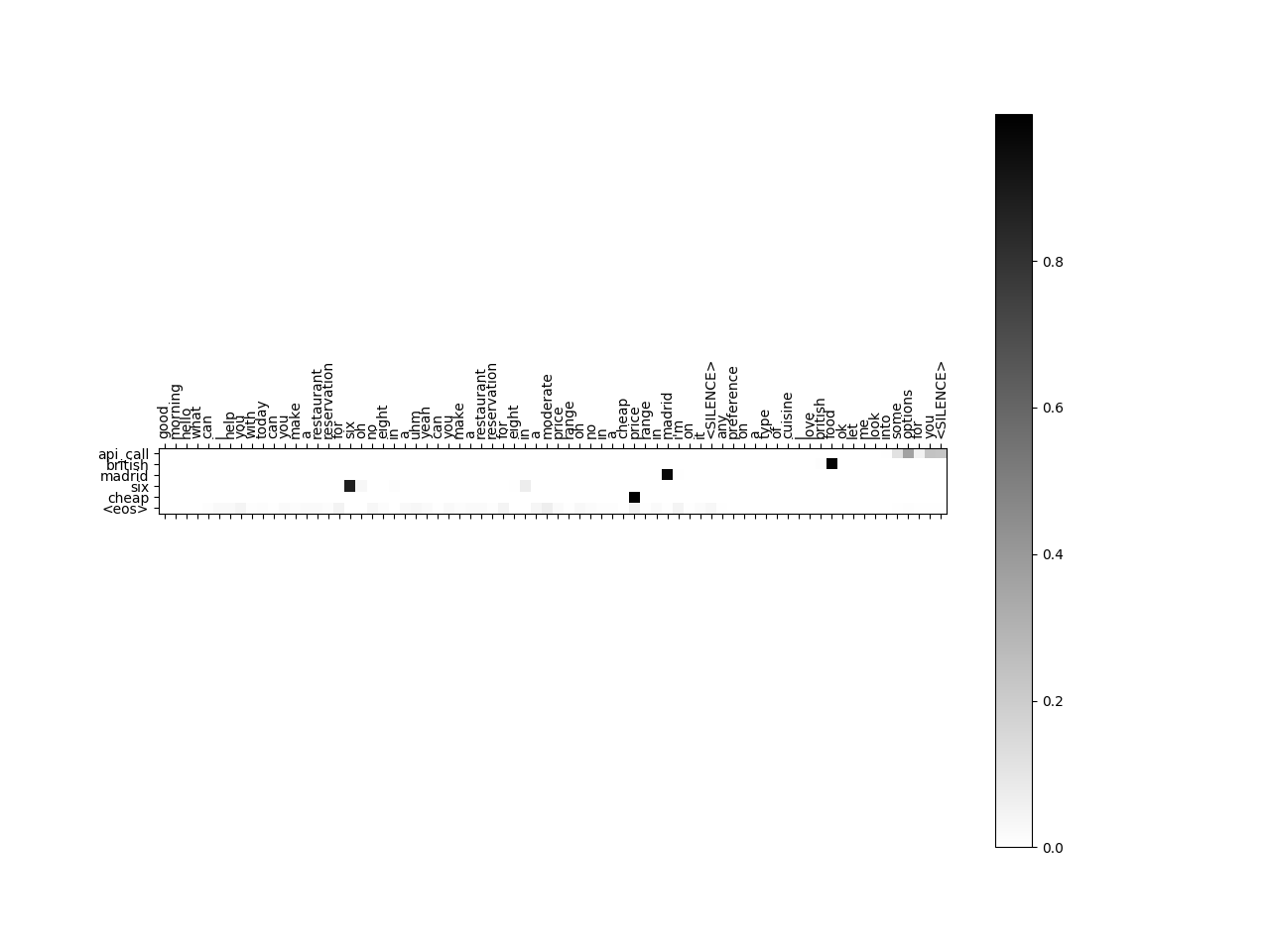}
\caption{Visualisation of the decoder attention when generating the API call (vertical axis) for the disfluent dialogue in example \ref{ex:repair} (horizontal axis). Darker colours indicate higher attention values.}\label{fig:attention}
\end{figure*}

We start by observing that the model faced with the bAbI/bAbI+ condition encounters new lexical items at test time, such as filled pauses (\texttt{uh}) or editing terms (\texttt{no sorry}).
These items are all mapped to a single token \texttt{<unk>} for `unknown'. 
In addition, the presence of disfluencies increases the length of user utterances: The average utterance length in bAbI is 4.8 tokens, while user utterances in bAbI+ have an average length of 7.6. 

Since the inventory of system utterances is very limited (there are 7 types of system outputs),
the model has no difficulty learning their surface form. The errors we find, therefore, do not concern the grammaticality of the utterances, but rather when in the course of a dialogue a particular system utterance is offered. We find that utterance errors always involve a jump ahead in the expected sequence of system utterances.
For example, around 83\% of the utterance errors occur at the beginning of the dialogue. If the first user utterance is disfluent  (e.g., \texttt{Good morning <unk> <unk> good morning}) and hence longer than expected, the model tends to skip the system's greeting and initial prompt (\texttt{Hello, what can I help you with today}) and instead outputs the second system prompt (\texttt{I'm on it}), which only makes sense once the user has already provided some task-related information. 
Utterance errors of this kind, where the model generates an utterance that would be appropriate later on in the dialogue, occur after all disfluency types present in bAbI+: hesitations, restarts, and self-repairs.

The current evaluation setup does not allow us to test whether utterance errors by the system lead to errors in the final API call, since the input sequence given to the encoder is the gold dialogue history and not the dialogue history with system utterances generated by the model. 
This, however, allows us to conclude that the errors made by the model on around 50\% of all API calls are a direct consequence of the disfluencies present in the user utterances. 
Not surprisingly, API call errors are triggered by self-corrections, which in bAbI+  always involve the mention of two different values for a single semantic slot (see example \ref{ex:repair} below). 
Indeed, the model trained on bAbI achieves near-perfect accuracy when tested on a corpus that does not contain self-correction but only hesitations and restarts. This indicates that the lower accuracy is specifically caused by the presence of multiple potential values for the same API slot in self-corrections.

Manual error inspection shows that the model does not follow a systematic strategy whereby the first mention of a semantic slot is selected, which would lead to all self-corrections being processed incorrectly. 
As illustrated in \ref{ex:repair}, the model seems to randomly select between the first and the second mention of a slot in a self-correction structure, which leads to some self-corrections being processed correctly: 

\ex. \label{ex:repair}
\noindent
\texttt{\small
\begin{tabular}[t]{@{}l@{}p{6.7cm}}
U: & good morning\\
S: & hello what can I help you with today\\
U: & can you make a reservation for six oh no eight in a uhm yeah can you make a reservation for eight in a moderate price range no in a cheap price range in madrid\\
S: & i'm on it\\
U: & <silence>\\
S: & any preference on a type of cuisine\\
U: & I love british food\\
S: & ok let me look into some options for you\\
U: & <silence>\\
S: & <API\_call british madrid six cheap>
\end{tabular}
}

\begin{figure*}
    \centering
        \includegraphics[width=0.35\linewidth, trim={15mm, 0mm, 21mm, 0mm}, clip]{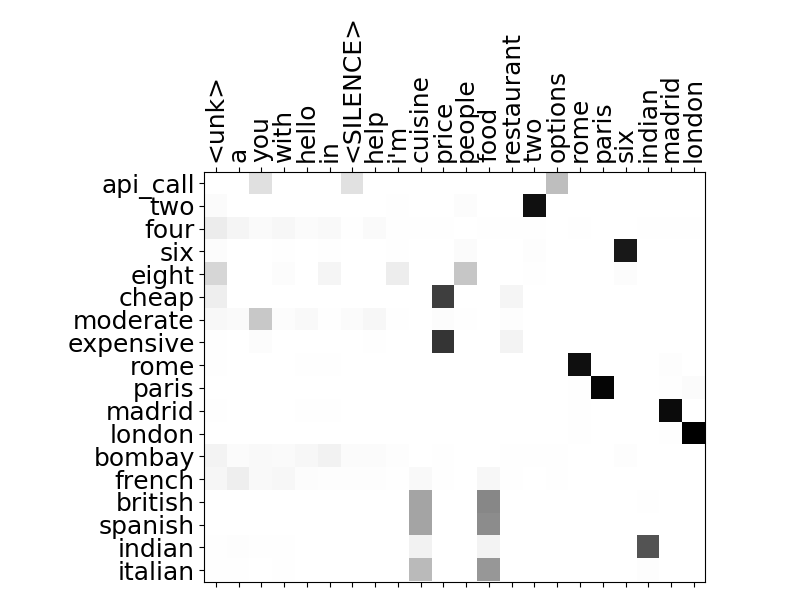}
        \includegraphics[width=0.35\linewidth, trim={15mm, 2mm, 35mm, 0mm}, clip]{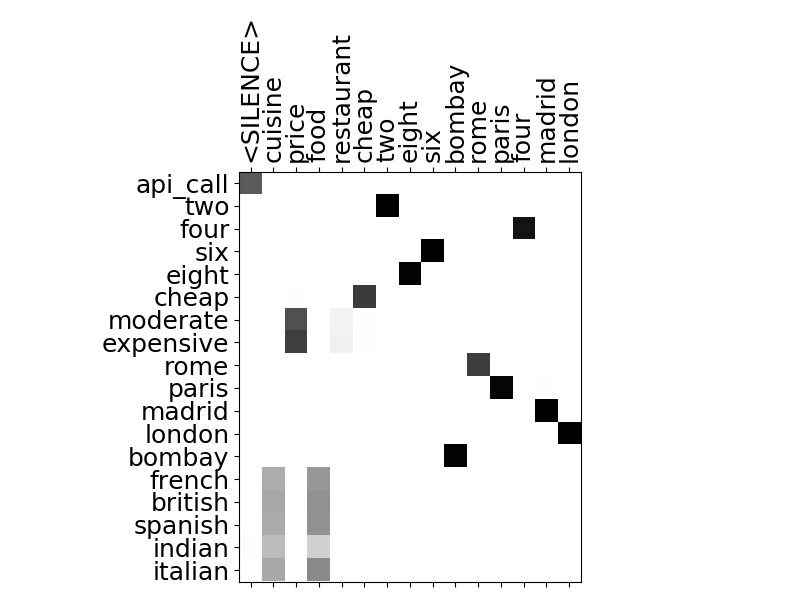}
    \caption{Alignment of in- and output words via the attention for different models tested on bAbI+ data. Left: model trained on bAbI. Right: model trained on bAbI+.}
 \label{fig:attention-alignment} 
\end{figure*}

\vspace*{3pt}
\noindent
In Figure \ref{fig:attention}, we visualise the decoder attention of the bAbI model when it generates the (partly incorrect) API call for the dialogue in \ref{ex:repair}. 
The plot shows that while generating the different components of the API call, the model strongly relies on its attention mechanism to select the right slot.
Furthermore, it confirms the observation that the model is not consistent in its strategy to select a slot after a disfluency: While it incorrectly attends to \texttt{six} (rather than \texttt{eight}), it correctly selects \texttt{cheap} (a repair of \texttt{moderate}).
In the next section, we will have a closer look at the attention patterns of both the bAbI and bAbI+ trained model.

%% file: attention.tex
\subsection{Comparing attention patterns}

To evaluate the network's attention patterns in a more quantitative way, we store all attention weights 
that the model computes while generating the API calls in the test corpus, and we compute their average for each API-call word.
That is, for all words in an API-call, we compute to which words in the dialogue history the decoder was on average attending while it generated that word.

We plot the results in Figure~\ref{fig:attention-alignment}, removing for each API-call word (on the vertical axis in the plot) the input words (horizontal axis) whose average attention score was lower 0.2.
We observe that the model trained on bAbI (left) not infrequently attends to \texttt{<unk>} while generating API calls, indicating that it did not attend to the correct content word. 
A striking difference appears comparing the results for the bAbI model with the bAbI+ trained model (Figure~\ref{fig:attention-alignment}, right), whose attention scores are much less diffuse.
While the bAbI model frequently attends to irrelevant words such as ``hello'', ``a'' or ``in'' (first columns in the plot), these words are not attended at all by the bAbI+ trained model.
This difference suggests that the bAbI+ model developed a more clear distinction between different types of words in the input and \textit{benefits} from the presence of disfluencies in the training data rather than being hindered by it.

In the next section, we investigate the representations developed by the bAbI+ model (bAbI+/bAbI+ condition), focussing in particular on how it incrementally processes disfluencies.

%% file: analysis2.tex
\section{Disfluency Processing}
\label{sec:dis-proc}

In contrast to previous work \cite{eshghi2017bootstrapping,shalyminov2017challenging}, our seq2seq model with attention trained on bAbI+ data learns to process disfluent user utterances remarkably well, achieving over 98\% sequence accuracy on API calls (see bAbI+/bAbI+ condition in Table~\ref{tab:results}). 
In this section, we investigate how the model deals with disfluencies, in particular self-corrections, by drawing inspiration from human disfluency processing. 

\subsection{The structure of disfluencies}

It has been often noted that disfluencies follow regular patterns \cite{levelt1983monitoring,shriberg:prelimdis}. Example \ref{ex:dis-structure} shows the structure of a prototypical self-correction, where the {\it reparandum} (RM) contains the problematic material to be repaired; the utterance is then interrupted, which can optionally be signalled with a filled paused and/or an {\em editing term} (ET); the final part is the {\em repair} (R) proper, after which the utterance may continue: 

\vspace*{5pt}
\ex. \label{ex:dis-structure}
\begin{tabular}[t]{@{}c@{\ }c@{\ }c@{\ }c@{\ }c}
a reservation  for  & {\em six}  & \{I mean\} & {\em eight} &  in a\ldots\\
                             &     RM       &      ET      & R & 
\end{tabular}

\vspace*{5pt}
\noindent
The presence or relationship between these elements serves to classify disfluencies into different types. For example, restarts such as those inserted in the bAbI+ corpus, are characterised by the fact that the reparandum and the repair components are identical (see example \ref{ex:restart} in Section~\ref{sec:babi+}); in contrast to self-corrections, where the repair differs from and is intended to overwrite the material in the reparandum. In hesitations such as \ref{ex:hesitation}, there is only a filled pause and no reparandum nor repair.

\subsection{Editing terms}
\label{sec:ET}

The algorithm used to generate the bAbI+ data systematically adds editing expressions (such as \texttt{oh no} or \texttt{sorry}) to all restarts and self-corrections inserted in the data. However, editing expressions (e.g., \emph{I mean, rather, that is, sorry, oops}) are in fact rare in naturally occurring human conversation. For example,  
\newcite{Hough15Thesis} finds that only 18.52\% of self-corrections in the Switchboard corpus contain an explicit editing term. 
Thus, while psycholinguistic research has shown that the presence of an explicit editing term followed by a correction makes the disfluency easier to handle \cite{brennan2001listeners}, humans are able to process disfluencies without the clues offered by such expressions. 

Here we test whether the model relies on the systematic presence of editing expressions in the bAbI+ data.  
To this end, we created two new versions of the dataset using the code by \citet{shalyminov2017challenging}:\footnote{\url{https://github.com/ishalyminov/babi_tools}} One with no editing term in any of the self-corrections or restarts, dubbed ``noET''; and one where there is an editing term in 20\% of self-corrections and restarts, dubbed ``realET'' as it reflects a more realistic presence of such expressions.
We refer to the original bAbI+ data, which has editing terms in all self-corrections and restarts, as ``fullET''. 

We test to what extent a model trained on fullET, which could rely on the systematic presence of an editing term to detect the presence of a self-correction or restart, is able to process disfluencies in a more natural scenario where editing expressions are only sparsely available (realET). 
The result indicates that the editing term has very little effect on the model's performance: as shown in Table~\ref{tab:edit}, accuracy goes down slightly, but is still extremely high (98\%).
This finding persists when the editing terms are left out of the test data entirely (97\% accuracy when testing on noET).
When models are trained on data containing fewer editing terms (realET and noET) and tested on data with a comparable or smaller percentage of editing terms, we observe a slightly larger drop in accuracy (see Table~\ref{tab:edit}).
We conclude that, although editing terms may help the model to develop better representations during training, their presence is not required to correctly process disfluencies at test time.

\begin{table}
\centering
\begin{tabular}{c|c|c|c|}
\hhline{~|---|}
& \multicolumn{3}{|c|}{\emph{Tested on}}\\
\hline
\multicolumn{1}{|c|}{\emph{Trained on}} & noET & realET & fullET  \\ \hline 
\multicolumn{1}{|c|}{fullET}            & 97 & 98   & 100    \\ \hline
\multicolumn{1}{|c|}{realET}            & 94 & 95             \\
\hhline{---|~|}
\multicolumn{1}{|c|}{noET}              & 94                   \\
\hhline{--|~~|}
\end{tabular}
\caption{Sequence accuracies of all sequences with and without editing term, averaged over 5 runs.}
\label{tab:edit}
\end{table}

\subsection{Identification of structural components}

Disfluencies have regular patterns. However, identifying their components online is not trivial. 
The comprehender faces what \newcite{levelt1983monitoring} calls the \emph{continuation problem}: the need to identify (the beginning and end of the reparandum and the repair onset. 
Evidence shows that there are no clues (prosodic or otherwise) present during the reparandum. Thus the identification of the disfluency takes place at or after the moment of interruption (typically during the repair). 
Here there may be prosodic changes, but such clues are usually absent \cite{levelt1983prosodic}. 
\newcite{FerreiraEtal2004} point out that ``the language comprehension system is able to identify a disfluency, likely through the use of a combination of cues (in some manner that is as yet not understood).''

We test to what extent our trained encoder-decoder model distinguishes reparanda and editing terms and can identify the boundaries of a repair using \textit{diagnostic classifiers} \citep{hupkes2018diagnostic}.
Diagnostic classifiers were proposed as a method to qualitatively evaluate whether specific information is encoded in high-dimensional representations---typically the hidden states that a trained neural network goes through while processing a sentence.
The technique relies on training simple neural meta-models to predict the information of interest from these representations and then uses the accuracy of the resulting classifiers as a proxy for the extent to which this information is encoded in the representations.

In our case, we aim to identify whether the hidden layer activations reflect if the model is currently processing a reparandum, an editing term, or the repair.
To test his, we label each word in the bAbI+ validation corpus according to which of the 3 categories it belongs to and train 3 binary classifiers to classify from the hidden layer activation of the encoder whether the word it just processed belongs to either one of these 3 classes.
For an example of such a labelling we refer to Figure~\ref{fig:dc_example}.

\begin{figure*}\small
    \centering
    \setlength{\tabcolsep}{2pt}
    \begin{tabular}{*{15}{c}}
        \textbf{\large \red 1} & \textbf{\large \blue 2} & \textbf{\large \blue 2} & \textbf{\large \green 3} & {\large 0} & {\large 0} & {\large 0} & \textbf{\large \red 1} & \textbf{\large \red 1} & \textbf{\large \blue 2} & \textbf{\large \blue 2} & \textbf{\large \green 3} & \textbf{\large \green 3} & {\large 0} & {\large 0} \\
        \texttt{with} & \texttt{uhm} & \texttt{yeah} & \texttt{with} & \texttt{british}& \texttt{cuisine} & \texttt{in} & \texttt{a} & \texttt{moderate} & \texttt{no} & \texttt{sorry} 
        & \texttt{a} & \texttt{cheap} & \texttt{price} & \texttt{range} 
    \end{tabular}
    \caption{A labelled example sentence to evaluate whether models have distinct representations for reparanda, repairs, and editing terms. 
    For each label, we train a separate binary classifier to predict whether or not a word belongs to the corresponding class.}\label{fig:dc_example}
\end{figure*}

We hypothesise that while reparanda will not be detectable in the hidden layer activations, as they can only be identified as such a posteriori \citep{levelt1983monitoring,FerreiraEtal2004}, editing terms should be easy to detect, since they belong to a class of distinct words.
The most interesting classifier we consider is the one identifying repairs, which requires a more structural understanding of the disfluency and the sentence as a whole.

\begin{table}[h]\centering
    \begin{tabular}{l|l@{\ }c@{\ }l|l@{\ }c@{\ }c|}\hhline{~|------|}
        & \multicolumn{3}{c|}{\textbf{self-corrections}} & \multicolumn{3}{c|}{\textbf{restarts}} \\
        \hline
 \multicolumn{1}{|l|}{   Reparandum}      & 15.0 & / & 89.4 & 27.4 & / & 92.6\\
 \multicolumn{1}{|l|}{   Editing term}    & 37.3 & / & 99.4  & 55.7 & / & 99.2\\
   \multicolumn{1}{|l|}{ Repair}         & 21.3 & / & 93.5 & 35.2 & / & 94.9 \\\hline

    \end{tabular}
    \caption{Precision / recall of diagnostic classifiers to identify reparanda, editing terms and repairs.}\label{tab:dc1}
\end{table}

\noindent
The general trends in our results (see Table~\ref{tab:dc1} above) are as expected: Editing terms are more easily recoverable than both reparanda and repairs, and the reparandum has the lowest scores with a precision and recall of 0.15 and 0.89, respectively. 
However, results for editing terms and repairs are lower than expected. The presence of editing terms is not reliably identifiable given the hidden layer activations of a model (37.3\% and 55.7\% precision for self-corrections and restarts, respectively), which is surprising given the fact that there is no overlap between editing terms and the rest of the model's vocabulary. 
Taken together with the results of our previous experiments in Section~\ref{sec:ET} regarding the effect of editing terms on the final sequence accuracy, this indicates that the presence of an editing term causes only minimal changes in the hidden layer activations, and thus leaves only a small trace in the hidden state of the network.
The performance of the repair classifier is also low: 21.3\% precision on self-correction and 35.2\% on restarts. These results suggest that the model has no explicit representations of the structure of disfluencies and instead relies on other cues to infer the right API call.

\subsection{Incremental interpretation}

Next we analyse how the model processes disfluencies by looking into the interpretation---in terms of task-related predictions---that the model builds incrementally, word by word and utterance by utterance.

\paragraph{Word by word}

First, we probe the representations of the \textit{encoder} part of the model while it processes incoming sentences, for which we use again diagnostic classifiers.
In particular, we test if information that is given at a particular point in the dialogue (for instance, the user expresses she would like to eat Indian food) is remembered by the encoder throughout the rest of the conversation.
We label the words in a dialogue according to what slot information was already provided previously in the dialogue, and test if this information can be predicted by a diagnostic classifier at later points in the dialogue.
Note that while in the bAbI data the prediction for a slot changes only once when the user expresses her preference, due to the possibilities of corrections, slot information may change multiple times in the bAbI+ corpus.
We train separate diagnostic classifiers for the different slots in the API call: cuisine (10 options), location (10 options), party size (4 options), and price range (3 options).

Our experiments show that the semantic information needed to issue an API call is not accurately predictable from the hidden representations that the encoder builds of the dialogue---see Table~\ref{tab:dc_slots}, where accuracy scores are all relatively low.

\begin{table}[h]
    \centering
\begin{tabular}{|l|c|} \hline
 Cuisine & 31.3\\
 Location & 25.9\\
 Price range & 57.3\\
 Party size & 	43.0\\ \hline
\end{tabular}
\caption{Accuracy per slot type in the word-by-word experiment.}\label{tab:dc_slots}
\end{table}

\begin{figure}\hspace*{-.3cm}
    \centering
\includegraphics[width=.48\textwidth, trim=5mm 0mm 15mm 14mm, clip]{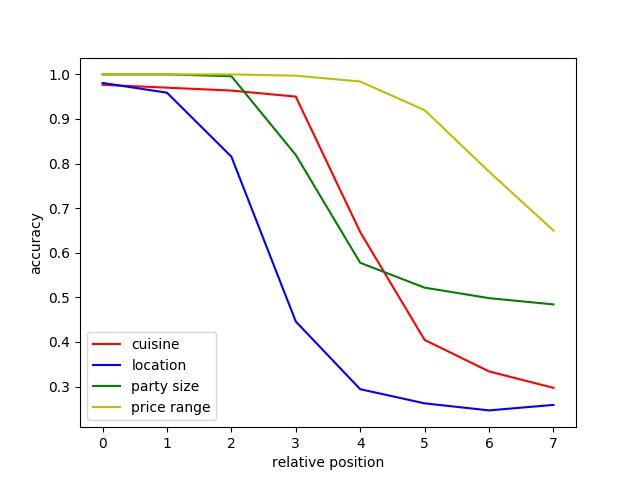}
\vspace*{-5pt}
\caption{Accuracy at position relative to mention in the dialogue of each type of slot.}
\label{fig:api_plot}
\end{figure}

\noindent
In Figure \ref{fig:api_plot}, we plot the accuracy of the diagnostic classifiers over time, relative to the position at which information appears in the dialogue (that is, the accuracy at position 4 represents the accuracy 4 words after the slot information occurred). 
The plot illustrates that the encoder keeps traces of semantic slot information for a few time steps after this information appears in the dialogue, but then rapidly `forgets' it when the dialogue continues.\footnote{To exclude the possibility that the low accuracy is a consequence of \textit{relocation} of information instead of it being forgotten, we also trained diagnostic classifiers to only start predicting a few words after slot information appears, but this did not result in an increase in accuracy.}
These results confirm our earlier findings that most of the burden for correctly issuing API calls falls on the model's attention mechanism, which needs to select the correct hidden states at the moment an API call should be generated.

\paragraph{Utterance by utterance}

In a second experiment, we study the incremental development of the API call made by the model's generative component (the {\em decoder}) by prompting it to generate an API call after every user utterance. 
To trigger the API calls, we append the utterances normally preceding an API call (\texttt{let me look some options for you <silence>}) to the dialogue history that is fed to the decoder.
We apply this trick to elicit an API call after every user utterance in the dialogue.
We evaluate the generated API calls by considering only the slots that can already be filled given the current dialogue history.
That is, in a dialogue in which the user has requested to eat Italian food in London but has not talked about party size, we exclude the party size slot from the evaluation, and evaluate only whether the generated API call correctly predicts ``Italian'' and ``London''.

For models trained on bAbI data, the described method reliably prompted an API call, while it was less successful for  models trained on bAbI+, where API calls were evoked only in 86\% of the time (when testing on bAbI+ data) and 54\% of the time (when testing on bAbI data). For our analysis, we consider only cases in which the API call was prompted and ignore cases in which other sentences were generated

\begin{table}
    \centering
    \begin{tabular}{|l@{\ }c@{\ }l|c|c|}
        \hline
        bAbI & / & bAbI   & 100 \\
        bAbI+ & / & bAbI  & 100 \\
        \hline
        bAbI & / & bAbI+  & 66.6\\
        bAbI+ & / & bAbI+ & 99.8\\
        \hline
    \end{tabular}
    \caption{Accuracy on triggered API calls utterance by utterance.}\label{tab:trig_api}
\vspace*{-10pt}
\end{table}

As shown in Table~\ref{tab:trig_api}, we find that the decoders of both the bAbI and bAbI+ models are able to generate appropriate API calls immediately after slots are mentioned in the user utterance ($\sim$100\% accuracy in the bAbI/bAbI, bAbI+/bAbI, and  bAbI+/bAbI+ conditions).
However, when confronted with disfluencies, the model trained on the disfluency-free bAbI data is not able to do so reliably (66.6\% accuracy with bAbI/bAbI+), following the trend we also observed in Table~\ref{tab:results}.

%% file: discussion.tex
\section{Conclusions}

We have investigated how recurrent neural networks trained in a synthetic dataset of task-oriented English dialogues process disfluencies. Our first conclusion is that, contrary to earlier findings, recurrent networks with attention can learn to correctly process disfluencies, provided they were presented to them at training time. 
In the current data, they do so without strongly relying on the presence of editing terms or identifying the repair component of disfluent structures.
When comparing models trained on data with and without disfluencies, we observe that the attention patterns of the former models are more clear-cut, suggesting that the disfluencies contribute to a better understanding of the input, rather than hindering it.

Furthermore, we find that in an encoder-decoder model with attention, at least for the current task-oriented setting, a large burden of the processing falls on the \textit{generative} part of the model: the decoder aided by the attention mechanism.
The encoder, on the other hand, does not incrementally develop complex representations of the dialogue history, limiting its usefulness as a cognitive model of language interpretation.
We preliminary conclude that different learning biases are necessary to obtain a more balanced division of labour between encoder and decoder.

Here we have exploited synthetic data, taking advantage of the control this affords regarding types and frequency of disfluency patterns, as well as the direct connection between language processing and task success present in the dataset. In the future, we aim at investigating neural models of disfluency processing applied to more naturalistic data, possibly leveraging eye-tracking information to ground language comprehension \cite{heller2015inferring,lowder2016prediction}.